\pdfoutput=1

\documentclass[11pt]{article}

\usepackage[preprint]{acl}

\usepackage{times}
\usepackage{latexsym}

\usepackage[T1]{fontenc}

\usepackage[utf8]{inputenc}

\usepackage{microtype}

\usepackage{inconsolata}

\usepackage{booktabs}
\usepackage{array}
\usepackage{multirow}

\usepackage{tcolorbox}
\usepackage{listings}
\usepackage{graphicx}
\usepackage{xcolor}
\usepackage{caption}
\usepackage{subcaption}
\usepackage{fdsymbol}
\usepackage{layouts}

\newcommand{\system}{Collage}
\newcommand{\dockerlogo}{
    \begingroup\normalfont
    \includegraphics[height=1em]{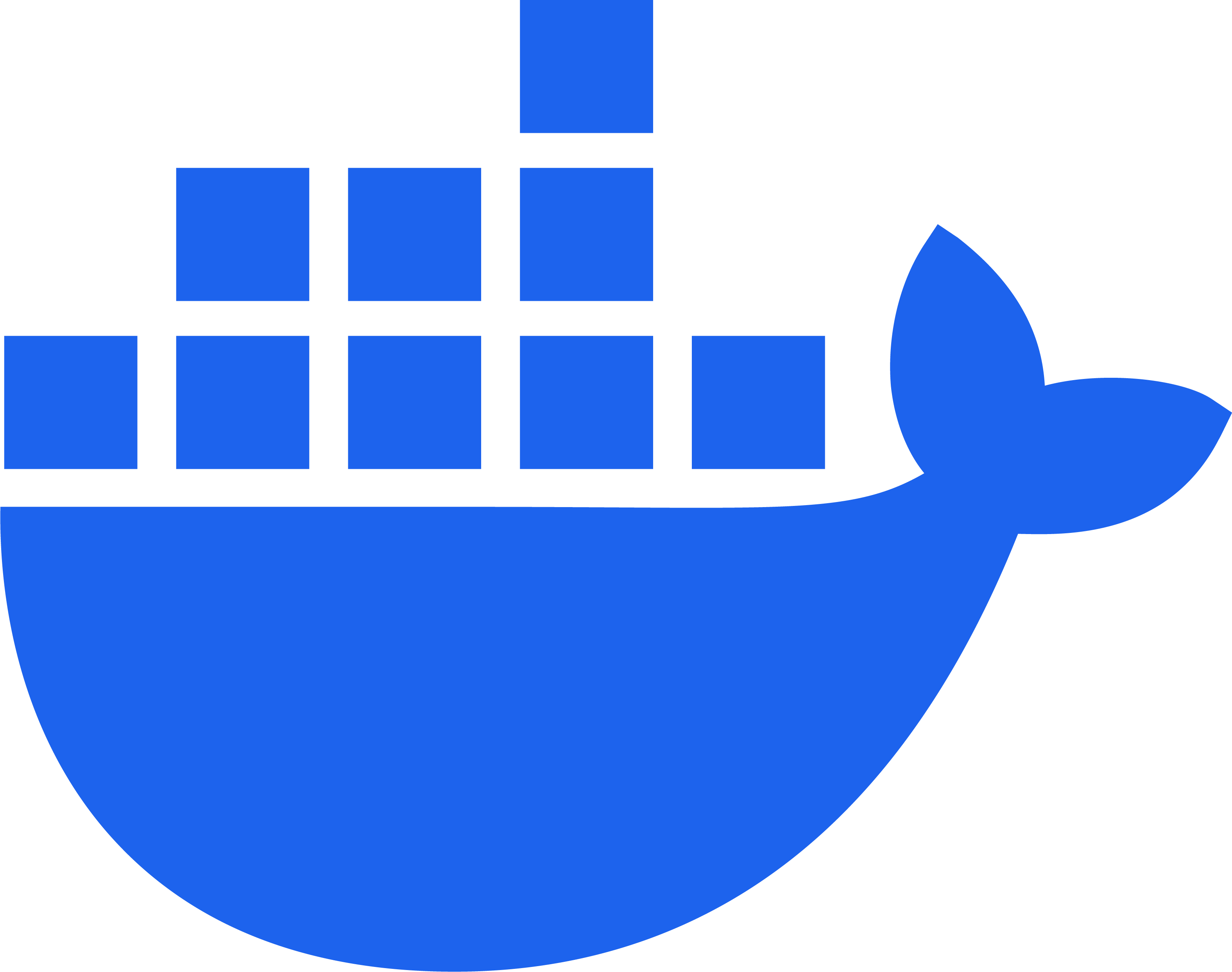}
    \endgroup
}
\newcommand{\runningcloud}{
    \begingroup\normalfont
    \includegraphics[height=1em]{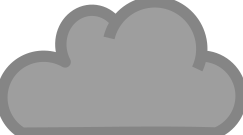}
    \endgroup
}

\usepackage{xcolor}
\definecolor{fig_gray}{HTML}{858585}
\definecolor{fig_blue}{HTML}{007cff}
\definecolor{fig_red}{HTML}{ff4845}
\definecolor{fig_yellow}{HTML}{ffd700}

\definecolor{fig_darkred}{HTML}{c81723}
\definecolor{fig_purple}{HTML}{811696}
\definecolor{fig_green}{HTML}{006228}

\title{\system{}: Decomposable Rapid Prototyping for Co-Designed Information Extraction on Scientific PDFs}

\author{
Sireesh Gururaja\textsuperscript{1}\thanks{\hspace{1mm}Equal contribution.}\ 
\quad Yueheng Zhang\textsuperscript{2}\footnotemark[1] \\  
\textbf{Guannan Tang}\textsuperscript{2}
\textbf{Tianhao Zhang}\textsuperscript{2} 
\quad \textbf{Kevin Murphy}\textsuperscript{2} 
\quad \textbf{Yu-Tsen Yi}\textsuperscript{2} 
\quad \textbf{Junwon Seo}\textsuperscript{2} \\ 
\textbf{Anthony Rollett}\textsuperscript{2} 
\quad \textbf{Emma Strubell}\textsuperscript{1,2} \\
\textsuperscript{1}Language Technologies Institute, School of Computer Science \\
\textsuperscript{2}Department of Materials Science and Engineering \\
    Carnegie Mellon University \\
    \texttt{sgururaj@cs.cmu.edu, yuehengz@andrew.cmu.edu}
}

\begin{document}
\maketitle
\begin{abstract}
Recent years in NLP have seen the continued development of domain-specific information extraction tools for scientific documents, alongside the release of increasingly multimodal pretrained language models. While applying and evaluating these new, general-purpose language model systems in specialized domains has never been easier, it remains difficult to compare them with models developed specifically for those domains, which tend to accept a narrower range of input formats, and are difficult to evaluate in the context of the original documents. Meanwhile, the general-purpose systems are often black-box and give little insight into preprocessing (like conversion to plain text or markdown) that can have significant downstream impact on their results.

In this work, we present \system{}, a tool intended to facilitate the co-design of information extraction systems on scientific PDFs between NLP developers and scientists by facilitating the rapid prototyping, visualization, and comparison of different information extraction models on the content of scientific PDFs. For scientists, Collage provides side-by-side visualization and comparison of multiple models of different input modalities in the context of the PDF content they are applied to; for developers, Collage allows the rapid deployment of new models by abstracting away PDF preprocessing and visualization into easily extensible software interfaces. We also enable both developers and scientists to inspect, debug, and better understand modeling pipelines by providing granular views of intermediate states of processing. We demonstrate our system in the context of information extraction to assist with literature review in materials science.

\end{abstract}

\section{Introduction}

In recent years, systems based on large language models (LLMs) have broadened the public visibility of developments in NLP. With the advent of tools that have publicly accessible, user-friendly interfaces, experts in specialized domains outside NLP are empowered to use and evaluate these models inside their domains, for example to automatically mine insights from scientific literature. Further, an increasing number of these tools are multimodal, handling not only text, but frequently images, or even PDFs directly. However, despite the accessibility of these tools, the processing pipelines they employ remain as end-to-end black boxes and provide little interpretability or debuggability in case of failure. Further, these systems usually rely only on  large, deployed models, potentially leaving other user priorities, such as interpretability, efficiency, or domain specialization, unaddressed.

\begin{figure}
    \centering
    \includegraphics[width=\linewidth]{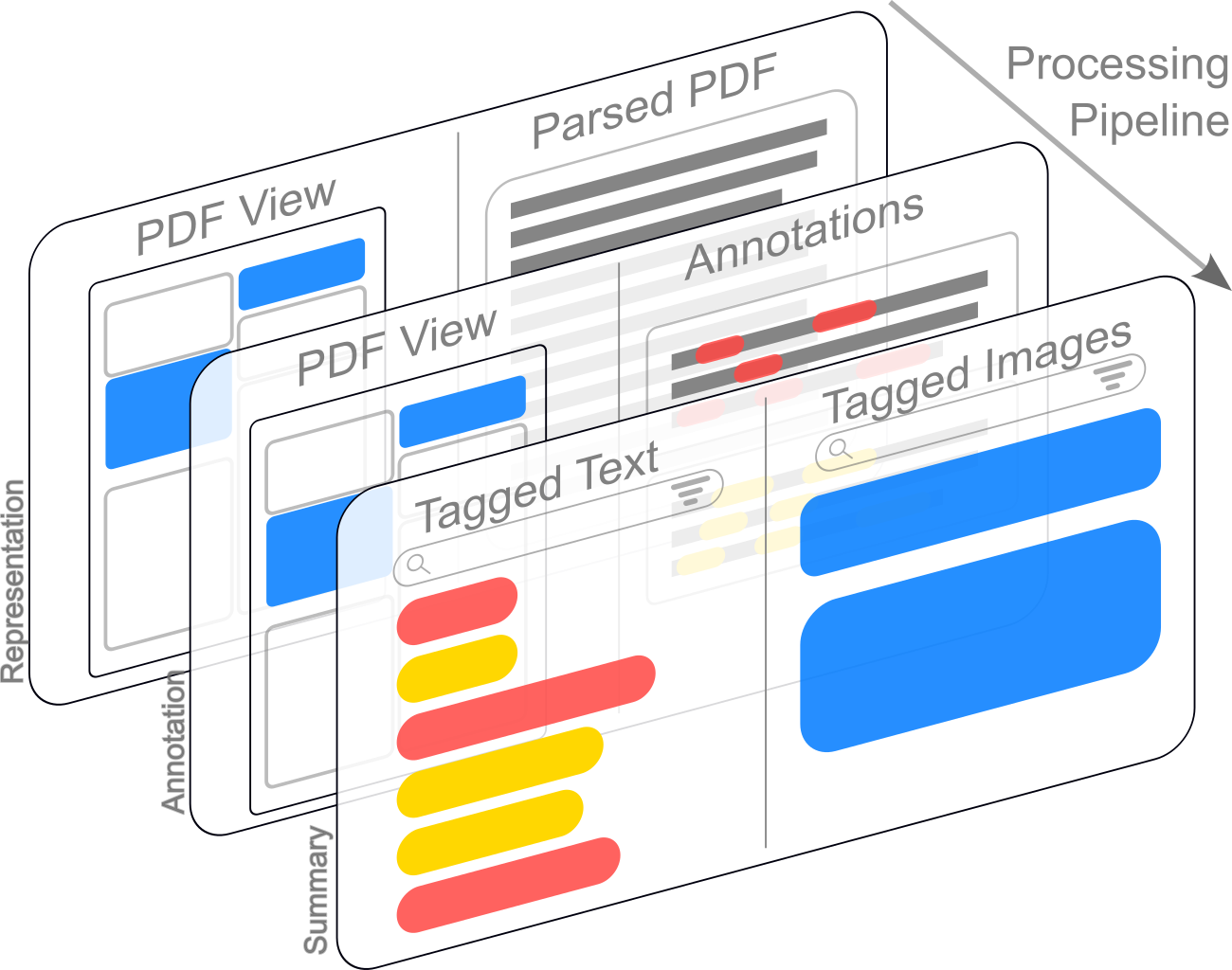}
    \caption{\system{} allows users to inspect multiple models in different modalities by presenting a stage-by-stage, decomposed view of the PDF modeling pipeline. Here, we see a PDF composed of \textcolor{fig_gray}{text} and \textcolor{fig_blue}{tables}, with entities from different models shown in \textcolor{fig_red}{red} and \textcolor{fig_yellow}{yellow}. The summary view shows extracted content, while annotations and inspection views allow the user to step back in the modeling pipeline}
    \label{fig:overview}
\end{figure}

Domain specific research in domains like clinical \citep{clinicalnlp-2023-clinical}, legal \citep{nllp-ws-2023-natural}, and scientific \citep{wosp-2020-international, sdp-2022-scholarly} NLP have long histories. Models in these areas remain less accessible; in order to run and evaluate these models on your own data, custom code is often needed. Further, because many of these models are text-only, evaluating their results in the context of their eventual use --- for example, directly on a PDF --- poses a challenge. 

This paper presents \system{}, a tool that facilitates the rapid prototyping, visualization, and comparison, of multiple models across modalities on the contents of scientific PDF documents. \system{} was designed to address the interface between developers of NLP-based tools for scientific documents and the scientists who are the intended users of those tools. To address scientists' needs, we ground our design in a series of interviews with domain experts in multiple fields, with a particular focus on materials science. Further, in cases where model results may not meet scientists' or developers' expectations, we visualize the intermediate representation at each step, giving the user a granular view of the modeling pipeline, allowing shared debugging processes between developers and users. \system{} is domain-agnostic, and can visualize any model that conforms to one of its three interfaces - for token classification models, text generation models, and image/text multimodal models. We provide implementations of these interfaces that allow the use of any HuggingFace token classifier, multiple LLMs, and several additional models without requiring users to write any code. All of the interfaces are easily implemented, and we provide instructions and reference implementations in our repository \footnote{\url{github.com/gsireesh/ht-max}}.

\begin{figure}[t!]
    \centering
    \includegraphics[width=\linewidth]{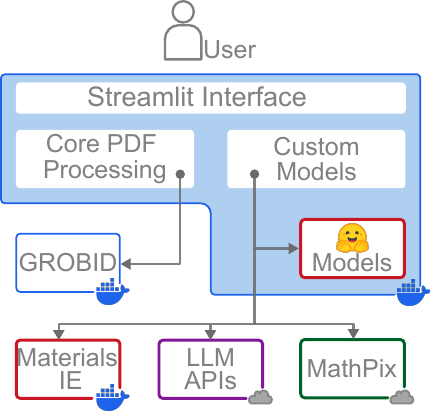}
    \caption{System architecture with currently implemented models. All custom models implement our interfaces, outline color indicates which: 
    \textcolor{fig_darkred}{Token Classification}, 
    \textcolor{fig_purple}{Text Generation}, or 
    \textcolor{fig_green}{Image Processing}. 
    \dockerlogo{} indicates components running in the same Docker container, and 
    \runningcloud{} indicates models running in the cloud. "Materials IE" refers to materials-specific models, like ChemDataExtractor.}
    \label{fig:system_arch}
\end{figure}

\section{Motivation}

\system{} is based on collected themes from interviews with 15 professionals across materials science, law, and policy, in which the authors ask about their practices for working with large collections of documents. For a reasonable scope, we focus on the 9 materials scientists in our sample, whose responses concern their process of literature review. We focus on three themes that emerged consistently from these interviews to inform our design of \system{}: 

\paragraph{Varied focuses.} One of the most prominent themes to emerge in our interviews is the variety of focuses that scientists, even in very closely related subfields, can have when reading a paper and evaluating it for relevance to their purpose. While many participants focused on paper  metadata, such as the reputation of the publication venue or citation count, others focused on cues from within the content of the paper. For the design of \system{}, we focus on accelerating co-design of models that address specific information extraction needs on paper content, by reducing the burden of deploying new models on PDF content, and providing a shared, user-friendly view of the results upon which scientists and developers can base subsequent efforts.

\paragraph{Information in tables.} As pointed out above, many of our participants relied heavily on information provided in tables, rather than solely in the document text. As such, an important concern in the design of \system{} would be to allow multimodality in the models that it interfaces with and visualizes.

\paragraph{Older documents.} Our participants noted that they regularly work with documents across a wide time range. Several participants noted that the work that they relied on most frequently were technical reports from the 1950s to the 1970s. These reports are now digitized, but are otherwise highly variable in their accessibility to modern processing tools: The OCR used when digitizing them can be inaccurate, they often contain noise in the scanned images, and layouts are less standardized. This can lead to confusion on whether issues with performance are the fault of models themselves, or preprocesing choices that cause that degraded performance. We therefore aim to provide an interface that allows users to inspect intermediate stages of processing, to better understand where a model may have failed, and what subsequent development should target next: whether better performing models, or better preprocessing.

\section{Design and Implementation}

We conceptualize our system in three parts: PDF representation, which parses and makes the content of PDFs easily available to downstream usage; modeling, i.e. applying multiple models to that PDF representation, backed by common software interfaces, which facilitate the rapid extension of the set of available models; and a frontend graphical interface that allows users to visualize and compare the results of those models on uploaded PDFs. We discuss the design choices  and implementation details of each stage in the following subsections, and show an architectural overview in Figure~\ref{fig:system_arch}.

\subsection{PDF Representation}
To produce a PDF representation amenable to our later processing, we build a pipeline on top of the PaperMage library~\citep{lo2023papermage}, which provides a convenient set of abstractions for handling multimodal PDF content. PaperMage allows the definition of \texttt{Recipe}s, i.e. combinations of processing steps that can be reused. We base our pipeline off of its \texttt{CoreRecipe} pipeline, which identifies visual and textual elements of a paper, such as tables and paragraphs. 

We then introduce several new components to the \texttt{CoreRecipe}, to make the paper representation more suitable to our use case. First, we introduce a parser based on Grobid \citep{GROBID}, which provides a semantic grouping of paragraphs into structural units, allowing us to segment processing and results by paper section. Second, to address issues with text segmentation in scientific documents, we replace PaperMage's default segmenter (based on PySBD) with a SciBERT \cite{beltagy2019scibert}-based SciSpaCy \cite{neumann-etal-2019-scispacy} pipeline. 

At the end of this stage of processing, we have the PaperMage representation of a document, in the form of \texttt{Entity} objects, organized in \texttt{Layer}s. \texttt{Entity} objects can be e.g. individual paragraphs by section or index, images of tables, and individual sentences. 

\subsection{Modeling and Software Interfaces}

To facilitate the easy implementation of new information extraction tools, we define common interfaces that simplify the process of adding additional processing to a document's content. These interfaces standardize three kinds of annotation on PDF content, allowing users convenient access to the PDF's content as images or strings (though they can access the PaperMage representation) and automatically handling visualization in several supported formats. This requires users to implement only a few simple functions in the modalities their models already use. All models currently in \system{} are implementations of these interfaces. We describe the interfaces, the requirements for implementation, and current implementations below. All interfaces are defined in the \texttt{papermage\_components/interfaces} package of our repository. In order to add a new custom processor, users must define a class that extends one of the interfaces specified below, and then register their predictor in the \texttt{local\_model\_config.py} module. 

\begin{figure}[t!]
\begin{lstlisting}[language=Python,rulecolor=\color{black},frame=single, keywordstyle=\color{magenta}]
class LiteLlmCompletionPredictor(TextGenerationPredictorABC):
  def __init__(
    self,
    model_name: str,
    api_key: str,
    prompt_generator_function: Callable[[str], List[LLMMessage]],
    entity_to_process="reading_order_sections",
  ):
    super().__init__(entity_to_process)
    self.model_name = model_name
    self.api_key = api_key
    self.generate_prompt = prompt_generator_function
  
  def generate_from_entity_text(self, entity: Entity) -> str:
    messages = [asdict(m) for m in self.generate_prompt(entity.text)]
    llm_response = completion(
      model=self.model_name, api_key=self.api_key, messages=messages, max_tokens=2500
    )
    response_text = llm_response.choices[0].message.content
    return response_text
\end{lstlisting}
    \caption{Partial implementation of the \texttt{TextGenerationPredictor} to allow LLM predictions given an \texttt{Entity} extracted from the PDF. \texttt{LLMMessage} is a data class wrapper around the system and user messages for LLMs in the OpenAI format. Not shown are the property declarations; full listing can be found in our \href{https://github.com/gsireesh/ht-max/blob/1c9f7dd76fecb73982a495866bc7f99cebd0b276/papermage_components/llm_completion_predictor.py}{code repository}.}
    \label{fig:llm_interface}
\end{figure}

\paragraph{Token Classification Interface:} This interface is intended for any model that produces annotations of spans in text, i.e. most ``classical'' NER or event extraction models. Users are required to extend the \texttt{TokenClassificationPredictorABC} class and override the \texttt{tag\_entities\_in\_batch} method, which takes a list of strings to tag, and produces a list of lists of tagged entities per-sentence. Tagged entities are expected to have the start and end character offsets, and the interface's code automatically handles mapping indices from the sentence level to the document level, and visualizing annotated text using the displaCy visualizer \footnote{\url{https://demos.explosion.ai/displacy-ent}}.

To demonstrate this interface, we provide two implementations: one with a common materials information extraction system, ChemDataExtractor2 \citep{swain_chemdataextractor_2016, mavracic2021chemdataextractor}, which we wrap in a simple REST API and Dockerize to streamline environment and setup, as well as a predictor that can apply any HuggingFace model that conforms to the \texttt{TokenClassification} task on the HuggingFace Hub\footnote{Model list available \href{https://huggingface.co/models?pipeline_tag=token-classification&library=transformers}{here}.}.

\begin{figure}
    \centering
    \includegraphics[width=\linewidth]{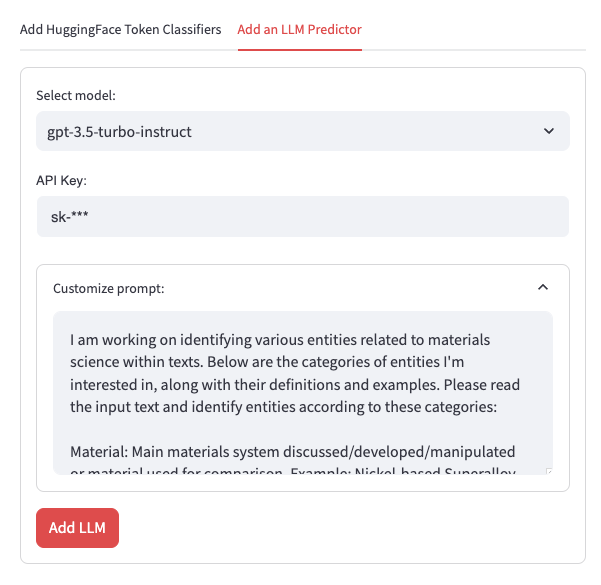}
    \caption{LLM Selector, as it appears in the File Upload view. Users specify an LLM to query, enter their API key, customize the prompt for an LLM, and repeat for any number of LLMs and prompts.
    }
    \label{fig:llm_selector}
\end{figure}

\paragraph{Text Generation Interface:} Given the prominence of large language model-based approaches, this interface is designed to allow for text-to-text prediction. Users are required to extend the \texttt{TextGenerationPredictorABC} class, and to implement the \texttt{generate\_from\_entity\_text()} method, which takes and returns a string. This basic setup allows users to e.g. prompt an LLM and display the raw response. A popular prompting method, however, is to request structured data e.g. in the form of JSON. To accommodate this, and to allow for aggregating LLM predictions into a table, users can also implement the \texttt{postprocess\_text\_to\_dict()} method. The default implementation of this method attempts to deserialize the entirety of the LLM response into a dictionary, but users can implement custom logic. 

Our implementation of this interface uses LiteLLM\footnote{\url{https://docs.litellm.ai/}}, a package that allows accessing multiple commercial LLM services behind the same API. We allow users to specify the endpoint/model, their own API key, and a prompt, and display predictions from that model. We show a partial implementation of this predictor in Figure~\ref{fig:llm_interface}, and a sample of its results in Figure~\ref{fig:annotation_view}.

\paragraph{Image Prediction Interface:} Given the focus on tables and charts that many of our interview participants discussed, and the fact that table parsing is an active research area, we additionally provide an interface for models that parse images, the \texttt{ImagePredictorABC} in order to handle multimodal processing, including tables. This interface allows users two options of method to override: In cases where only image inputs are needed (e.g. if a table extractor performs its own OCR), the \texttt{process\_image()} method; in cases where the method is inherently multimodal, implementors can instead override the \texttt{process\_entity()} method, which allows them full access to PaperMage's multimodal \texttt{Entity} representation. This interface requires implementors to return at least one of three types of data: a raw string representation, which we view as useful for e.g. image captioning tasks; a tabular dictionary representation, for the case of table parsing; or a list of bounding boxes, in the case of models that segment images. Implementations of this interface are free to return more than one type of output; all of them will be visualized in the frontend. 

We demonstrate implementations of both types. For raw image outputs, we implement a predictor that calls the MathPix API\footnote{\url{https://mathpix.com/}}, a commercial service for PDF understanding. For the multimodal approach, we implement a predictor that builds on the Microsoft Table Transformer model \citep{smock2023aligning}. This model predicts bounding boxes around table cells, which we then cross-reference with extracted PDF text in the PaperMage representation to provide parsed table output. An example of parsed table output from this predictor can be seen in figure \ref{fig:annotation_view}.

\subsection{Visualization Frontend}

\begin{figure*}[h!]
    \centering
    \includegraphics[width=6.3in, height=4in]{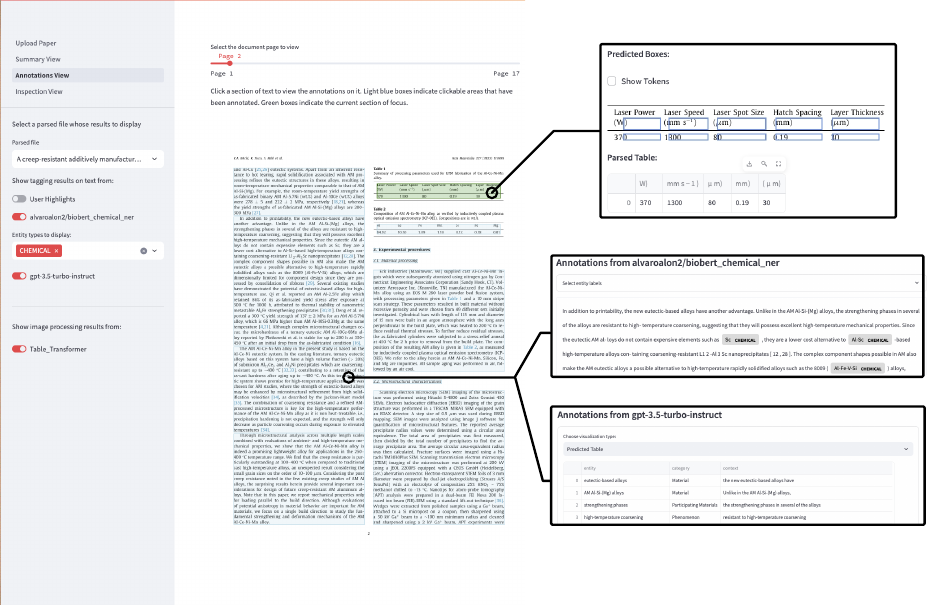}
    \caption{The annotations view. On the left, a screenshot showing the sidebar, allowing file and model selection, and the left pane, a visualization of the PDF with clickable regions highlighted. On the right, screenshots showing visualizations from the Table Transformer model with bounding boxes and parsed table (top), a HuggingFace transformer model with token-level tags (middle), and GPT-3.5 Turbo, with JSON output parsed into a table (bottom).
}
    \label{fig:annotation_view}
\end{figure*}

We present the results of the PDF processing in an interactive tool built using Streamlit\footnote{\url{https://streamlit.io}} that allows the user -- whether scientist or developer -- to upload a PDF, define a processing pipeline, and inspect the results of that processing pipeline at each stage. More concretely, after the paper is uploaded and processed, we present the results of the pipeline in three views, in decreasing order of abstraction from the paper. The intention of this is to first show the user the potential output of their chosen pipeline for a given paper, then allow them to inspect each step of the pipeline that led to that final output. Each view is described in more detail below, and has a screenshot in Appendix \ref{sec:appendix}.

\paragraph{File Upload and Processing.} The first view we present to a user allows them to upload a file, and to define the processing pipeline applied to that file. Basic PDF processing is always performed, and users can then toggle which custom models will be run. Users can additionally specify any number of HuggingFace token classification models or LLMs with the provided widget, which allows users to search the HuggingFace Hub, select LLMs, and customize the prompts for them. We show a view of the LLM model selector in Figure~\ref{fig:llm_selector}.

\paragraph{File Overview.} This view presents the high-level extracted information from the paper, as candidates for what could be shown to the user as part of their search process. In particular, we show a two-column view, with tables of tagged entities from both token-level predictors and LLMs on the left, and the processed content of images on the right. Users can filter based on sections, to e.g. find materials mentioned in the methods section of a paper. If the user finds the content extracted with the pipeline useful, the model and processing pipeline could be further developed into a more integrated prototype. If not, the user can proceed to the succeeding views, to see where models may have failed.

\paragraph{Annotations.} This view allows the user to compare the results of models in the context of the PDF. We present another two-column view, in which the PDF is visualized on the left, and allows the user to select a paragraph or table at a time, and visualize the results of each model on it. In the case of text annotation, we visualize the entities identified by token prediction models as well as predictions from LLMs. In the case of images, all of the available output types from the image processing interface are visualized. We show a composite screenshot of this interface in Figure~\ref{fig:annotation_view}.

\paragraph{Representation Inspection.} This view presents visualization of the PDF representation available to any downstream processing that the user might select. In the sidebar, users can choose to visualize any PaperMage \texttt{Layer}, i.e. set of  \texttt{Entity} objects, tagged by the basic processing steps. Then, in a view similar to the raw annotations view, they can see all of those entities highlighted on the PDF in the left-side column. Once the user selects an object, they see the raw content extracted from that object in the right-side column, in the form of its image representation and the text extracted from it, along with the option to view how the text is segmented into sentences. This view allows users to inspect how the PDF processing choices may have affected the text they send to models, which often have significant effects on their downstream performance \citep{camacho-collados-pilehvar-2018-role}.

\section{Addressing Needs from Interviews}
Our system is specifically designed to respond to the concerns raised in our interviews. First, to accommodate the varied processes of materials scientists, we design interfaces that allow for easy implementation of new models into our framework; our existing implementations of those interfaces also allow for the application of multiple LLMs and HuggingFace models directly in the context of the PDFs under review. This allows users to search for and evaluate models that suit their existing workflows. For tables, we both provide an interface and implementations that allow the comparison of proprietary and open-source table parsing systems. Extending this work to new table models and evaluating them is simplified by our software and visualization interfaces. Our inspection view is designed to address concerns about older PDFs: in being able to inspect the results of processing, users and engineers of this system can identify failure modes in both the upstream and downstream processing. 

\section{Co-design with Collage}

In this section, we walk through an example of how Collage might facilitate the development of an information extraction pipeline for a materials scientist. In this scenario, Bob, a materials scientist, wishes to extract the synthesis parameters of a class of materials called zeolites from a dataset of PDFs from the 1980s to the 2010s. Papers discussing Zeolite synthesis often report parameters both in the text of the paper as well as in tables, so multimodal extraction is crucial. He has worked with Alice, an NLP developer, before but they have not yet collaborated on this project. 

\paragraph{Evaluating off-the-shelf models.} Bob begins in Collage by trying to see if there is an existing model that already works for his case. Using the HuggingFace model selector in the upload paper view, he searches for tagging models, but only finds models trained on general scientific or biomedical text, not materials. He is, however, able to write a prompt for an LLM model to extract this information, and he adds predictors that call out to two popular commercial models to extract the information that he's interested in. He uploads a recent paper that he's been reading, and waits for Collage to process it. 

\paragraph{Finding modeling opportunities.} Once Collage has processed the paper, Bob heads to the \textbf{summary view}, and compares the results from the two commercial models. He's able to view the parameters that they extracted, filtered by section, to develop an understanding of what heuristics might get him the information he wants: parameters identified in the related works section, for example, are frequently irrelevant to his search. In the summary view, he's also able to see the tables that Collage has identified and parsed with the TableTransformer and MathPix models, along with their labels and captions, and the tagged bounding boxes for the table cells. 

To make sure those annotations are reasonable, he heads to the \textbf{annotations view}, where he can visualize the extracted information side-by-side with the original PDF content, and compare the annotations from his two LLMs. He's also able to check whether the table detection model has predicted sensible bounding boxes that both don't exclude content like table footnotes, but also don't include irrelevant, non-table content. He notes that while the table parsing from both models is reasonable, the paper he's reading reports values in ratios that may not be comparable across papers. To have a single pipeline that produces normalized results, he'd like to use a multimodal LLM, but in Collage currently, LLMs can only be applied to text. He decides to get in touch with Alice, to see if she can develop an LLM-based table information extraction model.

\paragraph{Prototype model development.} Alice begins work on a table information extraction tool, but there are a lot of possible options to evaluate: should she use a multimodal model and process the table in image format? Should she linearize the table into text, and have a text-only LLM work with it? In Collage, both options involve little more than implementing the LLM call, so it's easy to do both and then compare. For the multimodal case, Alice extends the \textbf{image predictor interface}, which allows her to receive as input the cropped image of any element on the page and pass that to an LLM; for the text-only case, Alice can easily access the underlying document representation use the already identified and parsed tables (which are in a DataFrame-compatible format) and convert them into markdown for her linearization. She is able to return a dictionary in the same schema for both predictors, which will automatically be visualized in the frontend as a Pandas dataframe. She commits her code, registers the predictors, and asks Bob to take a look in the Collage interface.

\paragraph{In-context evaluation.} Bob then re-processes his paper through Collage, making sure to check the boxes for Alice's new table parsing predictors. In the summary view, he's able to compare the predicted, normalized tables to the original PDF, to verify that the models are performing the normalization correctly. He then picks the better performing model, and asks Alice to create a pipeline that can process his entire dataset. Alice is able to take the predictor, add it to the PaperMage recipe that underlies Collage, and run it over Bob's set of PDF documents, adding a step to export the parsed tables that Bob saw in the Collage interface.

\paragraph{Diagnosing errors.}  Bob looks through the parsed tables from processing all of the PDFs, and notes that for the older PDFs, the parsed content doesn't look right. He'd like to diagnose the problem. Because the processing that Alice and Bob run on these documents is the same as that underlying Collage, the results can be visualized in the tool, even if they were not directly processed through it. Bob loads the representation of the parsed older document, and is able to view the results from the model that didn't look right. While the bounding boxes for the table look correct in the annotations view, he's also able to see in the \textbf{inspection view} that the text detected within the table has not been correctly OCR'd. He can now contact Alice to see if there's a fix for that problem, but in the meantime, he can examine the visualizations for his PDFs to understand how the publication year might affect whether the deployed suite of models can correctly extract and normalize information, and what the cutoff year might be for the results to be trustworthy. 

In this case, Collage enables Bob to self-serve cutting-edge NLP for his own use case, requiring that he involve Alice only when Collage's functionality needs extension. When that happens, Bob and Alice can both see results in the same interface, and can discuss errors and how to prioritize new work. When Alice develops new predictors to address Bob's needs, she is required to do no PDF processing or visualization, which are built into the tool, and Bob can evaluate and compare the results of these new predictors in the same interface he's been using the whole time. For debugging, both Bob and Alice have access to the same representation and visualization as a shared source of truth, and collaborate to involve both NLP and subject matter expertise in how to fix the problem. Collage can accelerate the process of collaboration between NLP developers and scientists, allowing for co-design and rapid prototyping with a shared representation.

\section{Related Work}

\system{} situates itself at the intersection of tools that offer reading assistance for scientific PDFs and tools that partially automate the process of literature review by means of information extraction. Tools for scientific PDFs often focuses on interfaces that augment the existing PDF with new information, such as citation contexts \citep{rachatasumrit_citeread_2022, nicholson_scite_2021}, or highlights that aid skimming \citep{fok_scim_2023}. However, most of these works are designed around and purpose-built for specific models. By contrast, \system{} draws from projects like PaperMage \citep{lo2023papermage}, by attempting to be model-agnostic, while at the same time providing a visual interface to prototype and evaluate those models. 

Scientific information extraction and literature review automation also have long histories. \system{}'s focus on materials science was driven by the field's existing investment into data-driven design \citep{himanen2019data, olivetti2020data}, which focuses on using information extraction tools to build up knowledge graphs to inform future materials research. This adds to the existing body of work in chemical and material information extraction, including works like ChemDataExtractor \citep{swain_chemdataextractor_2016, mavracic2021chemdataextractor} and MatSciBERT \citep{gupta2022matscibert}. Works like \citet{dagdelen_structured_2024} showcase the growing interest in LLM-based extraction; as LLMs increasingly become multimodal, this capability is likely to be used for tasks like scientific document understanding.
While all of these tools are intended to be applied to documents from the materials science domain, they do not share an interface: most tools expect plain text, some, like ChemDataExtractor allow HTML and XML documents, and some work with images. \system{} aims to be a platform on which multiple competing approaches can be evaluated, regardless of the input and output formats they require.


\section{Conclusion}
In this work, we present \system{}, a system designed to facilitate co-design and rapid prototyping of mixed modality information extraction on PDF content between scientists and NLP developers. We focus on a case study in the materials science domain, that allows materials scientists to evaluate models for their ability to assist in literature review. We intend for this work to be a platform on which to evaluate further modeling work in this area.

\section*{Ethics and Broader Impacts}
Our interview protocol was evaluated and approved by 
\makeatletter
\ifacl@anonymize
    our institution's IRB.
\else
    the Carnegie Mellon University Institutional Review Board as STUDY2023\_00000431.
\fi
\makeatother

In developing a tool to facilitate the automated processing of scientific PDFs, we feel that it is important to acknowledge that that automation may propagate the biases of the underlying models. Particularly in the case of English that does not reflect the training corpora that models were built on top of, models can perform poorly, leading to fewer results from those papers, and the potential to inadvertently exclude them. However, we hope that in providing a tool to inspect model outputs before such automation tools are deployed, that we can encourage critical evaluation and uses of these tools.

\section*{Acknowledgements}
\makeatletter
\ifacl@anonymize
    Anonymized for review.
\else
    Research was sponsored by the Army Research Laboratory and was accomplished under
    Cooperative Agreement Number W911NF-22-2-0121. The views and conclusions contained
    in this document are those of the authors and should not be interpreted as representing
    the official policies, either expressed or implied, of the Army Research Laboratory or the
    U.S. Government. The U.S. Government is authorized to reproduce and distribute reprints
    for Government purposes notwithstanding any copyright notation herein.
\fi
\makeatother

\bibliography{anthology, custom, zotero_references}

\appendix

\section{Appendix: Screenshots of Interface Views}
\label{sec:appendix}

\begin{figure*}
    \centering
    \includegraphics[height=4in]{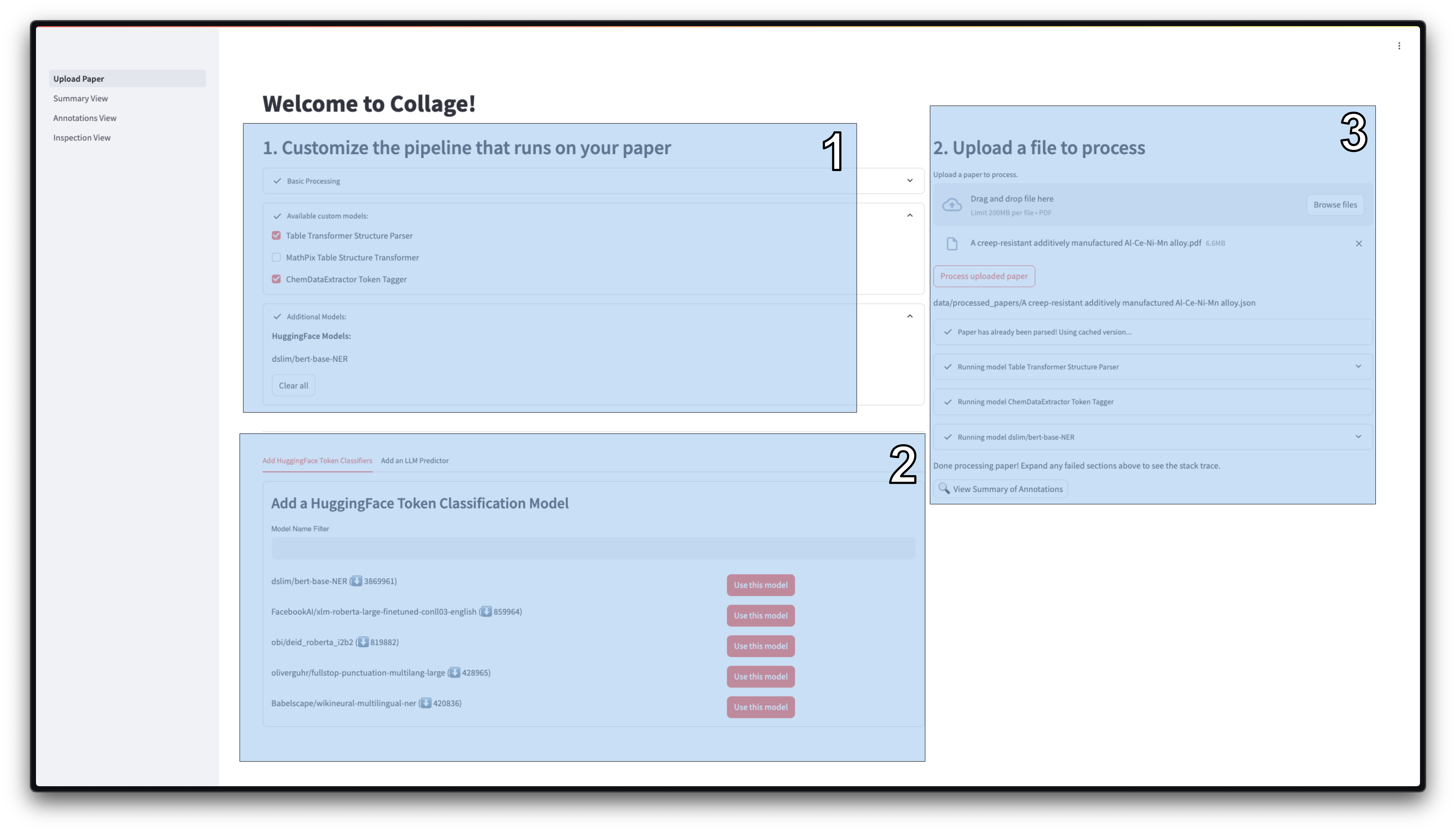}
    \caption{The Upload Paper view, showing (1) The currently selected models, (2) widget for selecting HuggingFace and LLM Classifiers, (3) File upload and progress visualization.}
    \label{fig:upload_view}
\end{figure*}

\begin{figure*}
    \centering
    \includegraphics[height=4in]{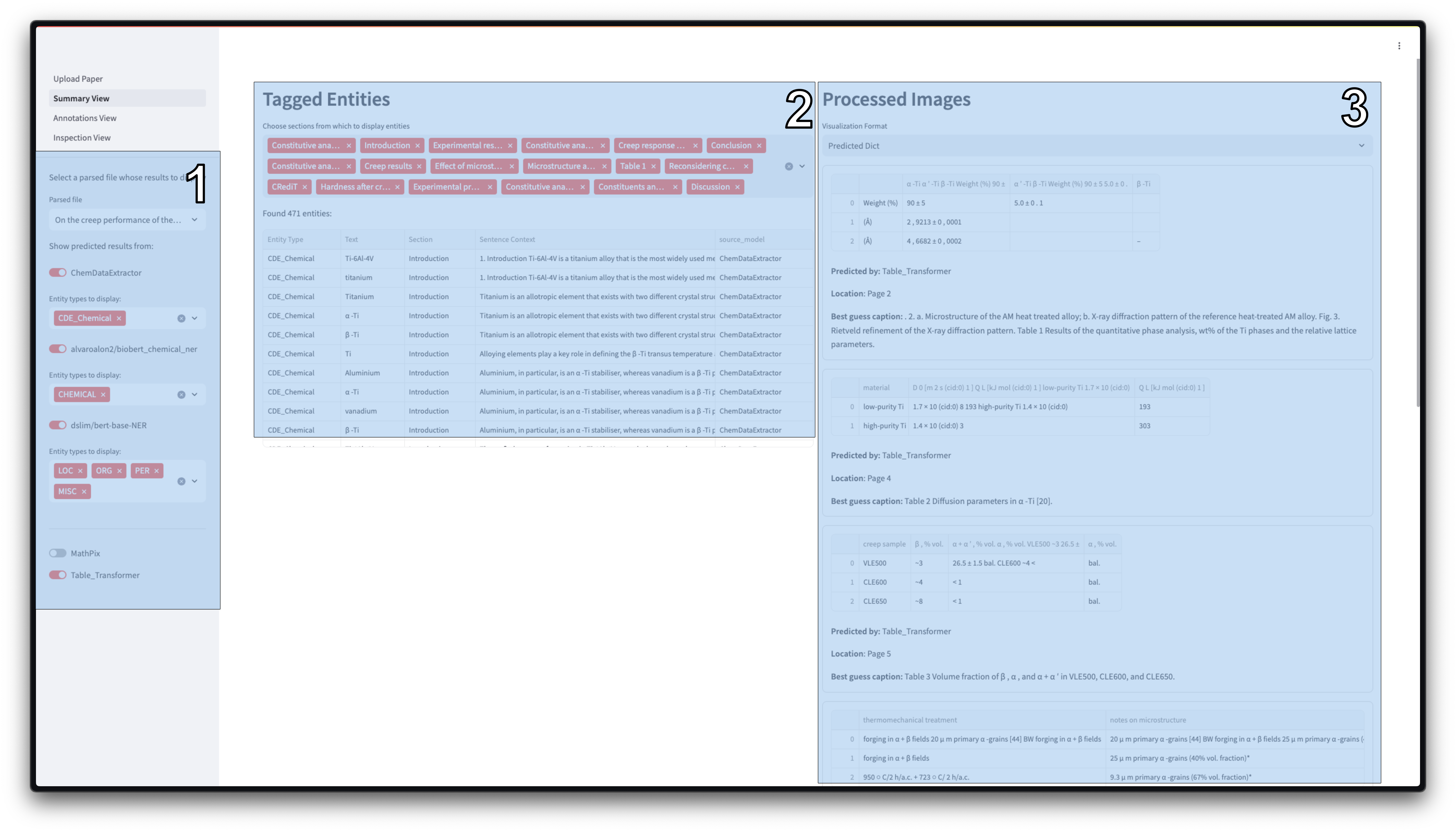}
    \caption{The Summary view, showing (1) the sidebar allowing model and entity type selection, (2) visualized tagged entities from the selected tagging models, (3) visualized image processing results.}
    \label{fig:summary_view}
\end{figure*}

\begin{figure*}
    \centering
    \includegraphics[height=4in]{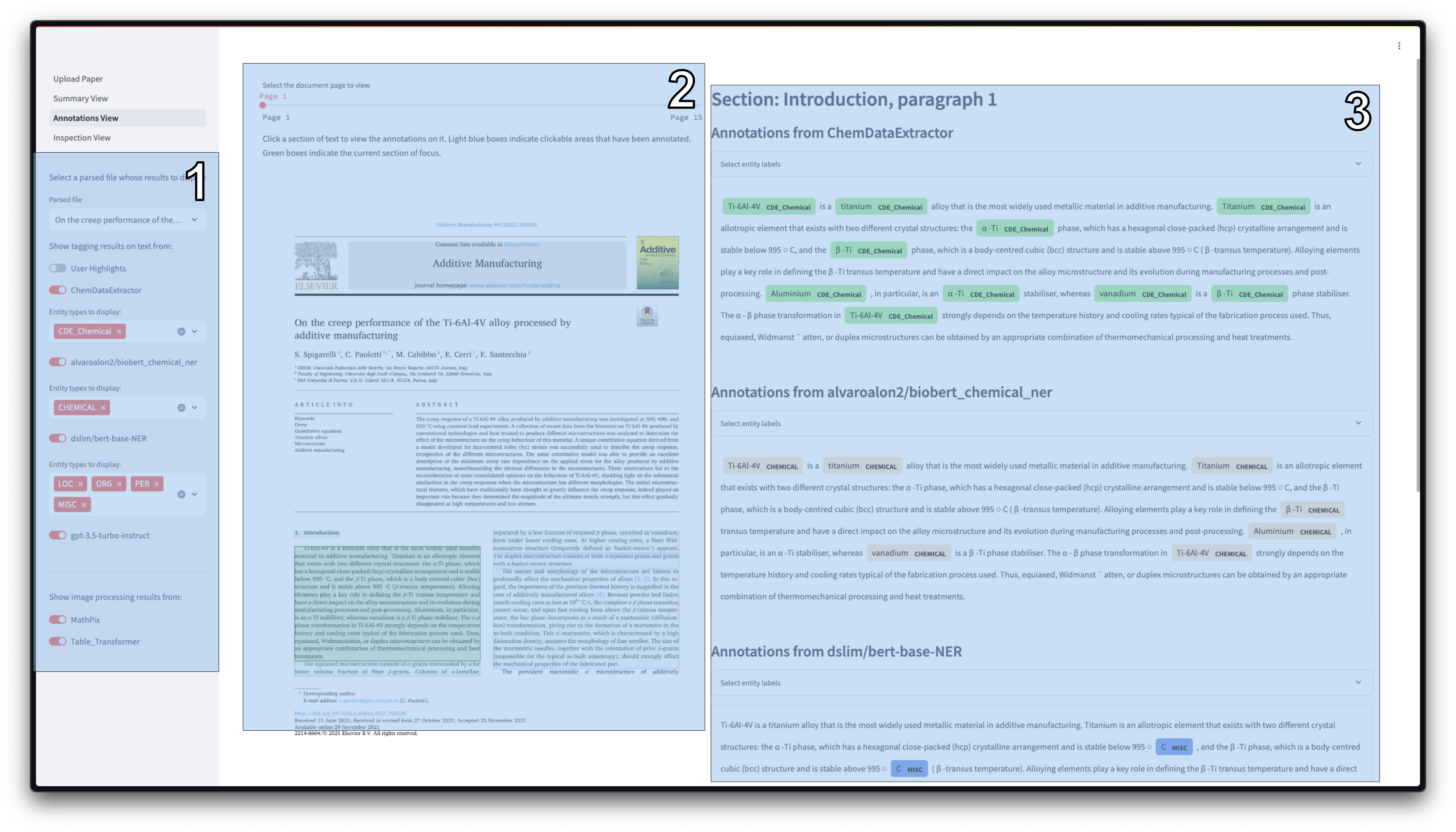}
    \caption{The Annotations view, showing (1) the sidebar allowing model and entity type selection, (2) the visualized PDF, showing clickable regions (3) visualized annotations on the clicked region.}
    \label{fig:annotations_view}
\end{figure*}

\begin{figure*}
    \centering
    \includegraphics[height=4in]{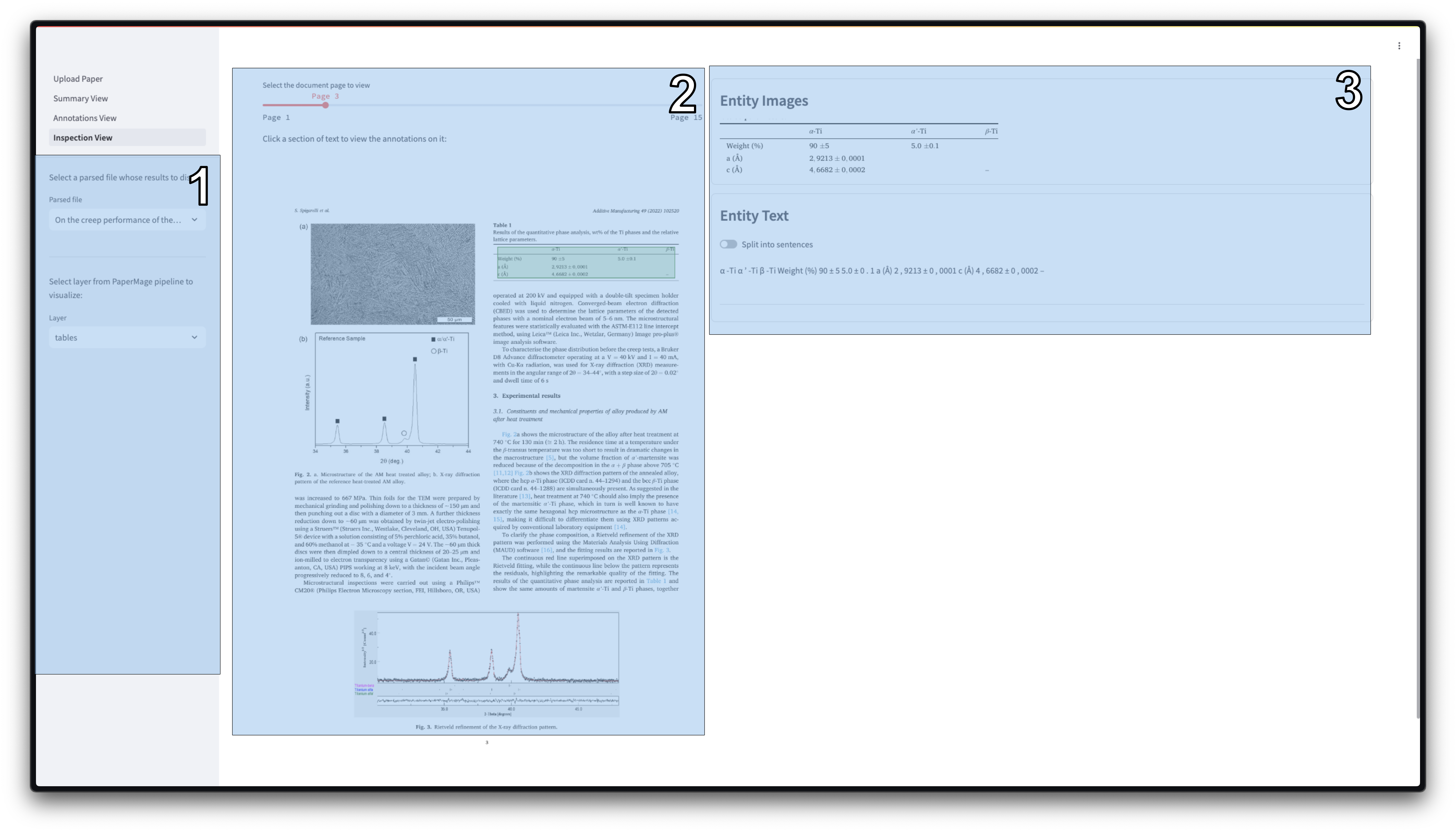}
    \caption{The Inspection view, showing (1) the sidebar allowing PaperMage layer selection, (2) the visualized PDF, showing clickable regions (3) the image and the text of the selected \texttt{Entity}}
    \label{fig:inspection_view}
\end{figure*}

\end{document}